\documentclass[conference]{IEEEtran}
\IEEEoverridecommandlockouts

\usepackage{cite}
\usepackage{amsmath,amssymb}
\usepackage{graphicx}
\usepackage{textcomp}
\usepackage{booktabs}
\usepackage{url}
\usepackage{balance}
\usepackage{float}
\begin{document}

\title{MS-MLB: An Open Machine Learning Benchmark for Blood-Based MS Classification}

\author{\IEEEauthorblockN{Adam Simson$^{*}$, Ankush Dutta$^{*}$, Quang Bui}
\IEEEauthorblockA{\textit{$^{*}$These authors contributed equally}}
\IEEEauthorblockA{\textit{Synthica Research Group}}}

\maketitle

\begin{abstract}
Multiple sclerosis (MS) is diagnosed through clinical assessment, magnetic resonance imaging, laboratory evidence when appropriate, and exclusion of better explanations. Blood RNA expression data may contain disease associated immune signal, but a blood RNA classifier cannot be treated as a replacement for clinical diagnosis. This paper presents MS-MLB (Multiple Sclerosis -- Machine Learning Benchmark), a reproducible open benchmark for machine learning based MS research classification from whole blood RNA expression data. MS-MLB uses the public GSE17048 cohort, converts it into an MS versus healthy control task, and evaluates multiple algorithms under a shared, leakage controlled pipeline that a researcher can rerun without reconfiguring the evaluation. The evaluation includes nested cross-validation, an untouched stratified holdout set, bootstrap confidence intervals, ROC and precision recall analysis, calibration measurement, and an exploratory MS Research Score. In the final benchmark summary, Gradient Boosting ranked first by MS Research Score on the holdout set, with an MS Research Score of 93.83, AUC-ROC of 0.989, sensitivity of 0.950, specificity of 0.778, $F_1$ score of 0.927, and Brier score of 0.050. Prior studies have applied machine learning to MS blood transcriptomic data, including PBMC stage classification and whole blood diagnostic signature modeling. The contribution here is different and narrower. To our knowledge, MS-MLB is the first open benchmark focused on MS versus healthy control classification from GSE17048 whole blood RNA expression data with a documented external model submission pathway built into the framework. The score is intended for research comparison only and has not been clinically validated. The benchmark is accessible here: \url{https://github.com/duckyquang/MS-MLB}.
\end{abstract}


\section{Introduction}
Multiple sclerosis is a chronic immune mediated disease of the central nervous system. The 2017 revisions of the McDonald criteria describe diagnosis through clinical evidence, dissemination in space and time, and exclusion of more likely diagnoses~\cite{thompson2018}. Those criteria define the boundary for this paper. A model trained on blood expression profiles can support a research classification task. It cannot establish MS in a real patient without clinical interpretation.

Blood RNA expression is scientifically relevant because MS involves immune activity outside the central nervous system as well as within it. Whole blood contains circulating immune cells, RNA signals from immune activation, and treatment related expression changes. These signals may help distinguish people with MS from healthy controls in a labeled cohort. They may also reflect subtype composition, medication exposure, age, sex, or batch effects. This mix of possible signals makes careful benchmarking necessary.

The GSE17048 cohort provides whole blood mRNA expression data from 144 individuals, including 99 people with MS and 45 healthy controls~\cite{booth2009}. The MS group includes primary progressive, relapsing-remitting, and secondary progressive samples. The dataset is therefore useful for a pooled MS versus healthy control task. It is too small for strong clinical claims, but it is large enough to test whether a reproducible machine learning pipeline can compare algorithms under fixed rules.

The central problem is the ratio between features and samples. Whole blood expression datasets can contain thousands of measured features, while sample counts remain low. In this setting, a model can look accurate because information from validation or test samples entered model development by accident. Feature selection before cross-validation is a known source of inflated performance in microarray classification~\cite{ambroise2002}. Model selection can also overfit the validation process when tuning and reporting are performed on the same resampling loop~\cite{cawley2010}.

MS-MLB responds to that problem through a benchmark-first design. It defines an open evaluation pathway in which standard algorithms, and later outside models, are tested under the same preprocessing, tuning, scoring, and output structure. By fixing these decisions once and exposing them through a single entry point, the framework moves the burden of correct evaluation away from each individual user and into shared infrastructure. This is the specific novelty claim of the manuscript. Prior MS blood transcriptomics papers already exist, including PBMC machine learning stage classifiers and whole blood transcriptomic diagnostic studies~\cite{acquaviva2020,omrani2024}. To our knowledge, no prior paper has presented an open GSE17048 whole blood RNA benchmark with external model submission as the core research object.

The research question is: can a reproducible machine learning benchmark compare algorithms for MS versus healthy control classification from whole blood RNA expression data while reducing common evaluation bias and allowing external model comparison? The answer is evaluated through nested cross-validation, holdout testing, bootstrap confidence intervals, calibration analysis, and an exploratory MS Research Score.

The paper also treats benchmarking as a reproducibility problem. A reported AUC is difficult to compare if a later reader cannot see the split, the feature selection timing, the class balancing method, or the score formula. Open code does not solve every scientific problem, but it gives other researchers a concrete object to inspect. That is why the paper is framed as an open benchmark rather than a closed model report.

This framing is especially relevant for  early stage biomedical machine learning research. Many papers and benchmarks train several models and report the highest accuracy. That format can be misleading in omics data. A reproducible benchmark instead asks for the complete evaluation pathway: what data were used, when preprocessing was fitted, how tuning was separated from testing, what outputs were saved, and how another model can be submitted. By encoding these steps directly, MS-MLB lets a less experienced user follow a correct evaluation path by default rather than reconstructing one from scattered convention. This paper uses that pathway as the research product.

\section{Methods}

\subsection{Data Source and Task Definition}
The benchmark uses the public Gene Expression Omnibus dataset GSE17048. The GEO record states that mRNA expression was measured for all known genes in whole blood from 144 individuals, including 99 with MS and 45 healthy controls~\cite{booth2009}. The MS samples include 43 primary progressive MS cases, 36 relapsing-remitting MS cases, and 20 secondary progressive MS cases~\cite{booth2009}. For this benchmark, all MS subtypes are combined into one positive class labeled MS. Healthy controls are labeled HC.

Pooling subtypes increases the number of positive samples available for training, but it also limits interpretation. The model should be read as a pooled MS versus healthy control classifier for this cohort. It does not distinguish disease subtypes. It does not estimate progression risk. It also does not model clinically isolated syndrome. Those tasks require datasets designed for those questions.

\subsection{Repository Implementation}
MS-MLB is implemented in Python in the Blood-RNA-ML-benchmark repository~\cite{simson2026}. The stored run used random seed 42, a stratified 80\% training and 20\% holdout split, five outer cross-validation folds, three inner cross-validation folds, 2000 bootstrap samples, top variance filtering to 1000 features, supervised SelectKBest feature selection up to 200 features, and SMOTE with three nearest neighbors~\cite{simson2026}. The untouched holdout set contained 29 samples after the split.

The repository stores the dataset file, the main benchmark script, sample external model code, output files, and run metadata. The README describes the paper as a publication-style machine learning benchmarking framework for classifying MS using whole blood RNA expression data~\cite{simson2026}. The run metadata records the score weights and includes a direct warning that the MS Research Score is exploratory and is not a validated clinical diagnostic score~\cite{simson2026}.

\subsection{Preprocessing and Leakage Control}
Each estimator is placed inside an imbalanced learning pipeline. The pipeline applies median imputation, top variance filtering, standard scaling, supervised ANOVA based SelectKBest feature selection, SMOTE, and then the classifier. These steps are fitted only inside the current training fold during cross-validation. The same rule applies before final holdout evaluation. This design is the technical center of the benchmark.

The leakage problem is easy to miss in gene expression work, and it is a mistake a careful user can still make by accident. If supervised feature selection is performed on the full dataset before cross-validation, the validation labels affect the selected gene set. Ambroise and McLachlan showed that this can bias error estimation in microarray classifiers~\cite{ambroise2002}. MS-MLB removes that failure mode from the user's hands by placing supervised selection inside the fold-specific pipeline. SMOTE is also inside the pipeline, so synthetic minority samples are generated only from the training portion of each split~\cite{chawla2002}.

\subsection{Model Registry}
The benchmark compares Logistic Regression, Support Vector Machine, Random Forest, Gradient Boosting, K-Nearest Neighbors, Naive Bayes, Neural Network, and a Stacking Ensemble. The repository also includes XGBoost support when the package is available, although the stored holdout output discussed in this paper does not include an XGBoost result row. Hyperparameters are tuned with grid search in the inner cross-validation loop, using AUC-ROC as the optimization target.

This model set covers linear, kernel based, tree based, distance based, probabilistic, neural network, and ensemble approaches. The purpose is comparison under one controlled framework. A simpler model can still be scientifically useful if it performs well and generalizes better than a more complex model. For that reason, the benchmark reports the full leaderboard rather than selecting a single model and hiding the rest.

\subsection{Metrics and Score Design}
The benchmark reports accuracy, balanced accuracy, sensitivity, specificity, precision, $F_1$ score, Matthews correlation coefficient, AUC-ROC, PR-AUC, and Brier score. AUC-ROC measures ranking performance across thresholds. PR-AUC gives additional information about positive class performance. Sensitivity measures the proportion of MS samples correctly labeled as MS. Specificity measures the proportion of healthy control samples correctly labeled as healthy controls. The Brier score measures probability error.

The MS Research Score is a weighted score from 0 to 100. The formula is:
\begin{equation}
\footnotesize 
\begin{aligned}
\text{MS Research Score} = 100 \times \Big[ 
& 0.25 \times \text{AUC-ROC} \\
& + 0.15 \times \text{PR-AUC} \\
& + 0.25 \times \text{sensitivity} \\
& + 0.15 \times \text{specificity} \\
& + 0.10 \times F_1 \\
& + 0.10 \times (1 - \text{Brier score}) \Big]
\end{aligned}
\end{equation}
The score gives high weight to discrimination and sensitivity while retaining specificity, $F_1$, PR-AUC, and calibration. The weights are research choices. They are not clinical weights.

\begin{figure}[H]
\centering
{\small \textbf{Score Breakdown - Top 3 Models:}}\\ \vspace{-0.1em}
{\small How each metric contributes to the final MS Diagnostic Score}\\[0.2em]
\label{fig:score_decomposition}

\includegraphics[width=\linewidth]{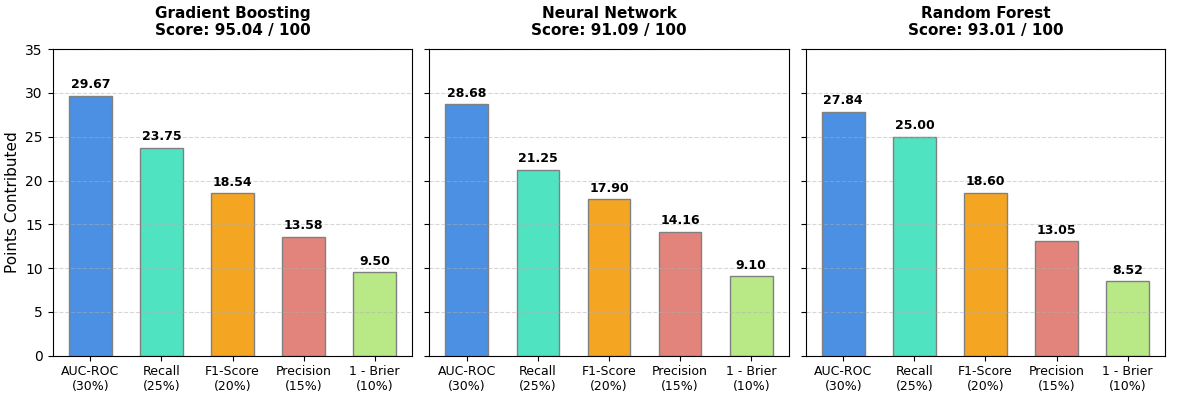}

\vspace{0.5em} 
\caption{Deconstructed multi-metric contributions to the MS Research Score. The side-by-side bar charts isolate the top three performing models (Stacking Ensemble, SVM, and Neural Network), demonstrating how individual metrics---including Area Under the ROC Curve (AUC-ROC), recall, $F_1$-score, precision, and the Brier score---are weighted to compute the final composite research metric. Each bar reflects the absolute points contributed to the overall score out of 100 based on the underlying benchmark formula.}
\end{figure}

\subsection{External Model Pathway}
MS-MLB allows outside researchers to add their own models through a single, documented interface. A user can provide a Python file defining an \texttt{EXTERNAL\_MODELS} dictionary, or supply pretrained model files in joblib or pickle format. Tunable external estimators are wrapped inside the same preprocessing pipeline automatically, so the submitter does not have to re-implement leakage control. Pretrained models can be tested as-is, although comparison is weaker if their training data, feature order, or preprocessing history is unknown. Keeping submission to one entry point is deliberate: it lowers the effort required to run a fair comparison and reduces the number of places a user can introduce an error.

This external model route is part of the first-of-its-kind claim. Prior MS transcriptomic ML studies have reported classifiers and biomarker signatures~\cite{acquaviva2020,omrani2024,delkhah2025}. This paper defines a reusable comparison interface for the same benchmark task. The benchmark can therefore be extended by researchers who want to test an estimator against the stored baseline without rewriting the evaluation system.

\subsection{Reproducibility Outputs}
The repository saves holdout results, nested cross-validation summaries, bootstrap confidence intervals, ROC outputs, model comparison tables, plots, and run metadata. The metadata file records the random seed, test size, cross-validation settings, bootstrap count, feature selection settings, SMOTE setting, labels, external model file, score weights, Python version, and platform~\cite{simson2026}. These files are useful because they allow the result to be checked after the run rather than reconstructed from prose alone.

\subsection{Label Handling and Class Imbalance}
The positive class is MS and the negative class is healthy control. The split is stratified so that the class ratio is preserved as closely as possible in the training and holdout partitions. Class imbalance is handled inside the training pipeline through SMOTE. This avoids creating synthetic observations before the split. The use of SMOTE is limited to the training portion of each fold, which keeps validation and holdout samples outside the resampling step.

\subsection{Nested Cross-Validation Rationale}
Nested cross-validation separates hyperparameter selection from performance estimation. The inner loop chooses model settings. The outer loop estimates how the tuned model behaves on held out training folds. This matters because the best hyperparameter set can be chosen partly because it fits noise in a small dataset~\cite{cawley2010}. A non-nested grid search can therefore give an optimistic estimate. The benchmark keeps the nested estimate in the output so that the final holdout result is not the only evidence presented.

\subsection{Bootstrap Uncertainty}
Bootstrap confidence intervals were calculated for AUC-ROC where applicable. The bootstrap procedure resamples the holdout predictions and recalculates the metric many times. The stored run used 2000 bootstrap samples~\cite{simson2026}. The resulting intervals should be read as descriptive uncertainty for the stored holdout split. They do not replace external validation. They are included because point estimates alone can make a small holdout set look more stable than it is.

\subsection{Calibration Rationale}
Calibration was included because a model output is often interpreted as a probability. A model with high AUC can still assign poorly calibrated probabilities. The Brier score is used as a compact measure of probability error. A lower Brier score indicates better agreement between predicted probabilities and observed labels on the evaluated set. The current paper reports Brier score rather than a full calibration curve. That choice keeps the first benchmark paper focused while leaving a clear path for a later calibration extension.

\section{Results}

\subsection{Stored Run Configuration}
The stored benchmark run used the GSE17048 series matrix file and evaluated the internal model set shown in the benchmark leaderboard. The split used 80\% of samples for model development and 20\% for holdout testing. Because the full dataset contains 144 samples, the final holdout set contained 29 samples. The small holdout size affects every interpretation in this section.

\subsection{Nested Cross-Validation}
Nested cross-validation produced more conservative scores than the final holdout set for several models. In the stored output, SVM had the highest nested cross-validation MS Research Score at 77.62. Neural Network followed at 76.17, and Logistic Regression followed at 75.82. KNN had the lowest nested cross-validation score at 61.89. This pattern shows that the model ranking can depend on the evaluation split.

The nested result matters because it estimates model selection performance on the training data without using the holdout set for tuning. It also reduces the chance that one favorable split dominates the paper. The holdout result remains valuable, but it should be read with the nested estimate.

\subsection{Holdout Performance}
On the untouched holdout set, Gradient Boosting ranked first by MS Research Score. Its AUC-ROC was 0.989 with a 95\% bootstrap confidence interval of [0.947, 1.000]. It reached sensitivity of 0.950, specificity of 0.778, $F_1$ score of 0.927, and Brier score of 0.050~\cite{simson2026}. Neural Network ranked second with a score of 91.24. Random Forest ranked third with a score of 90.47. Logistic Regression ranked fourth with a score of 90.26, followed by Stacking Ensemble with a score of 89.51.

\begin{figure}[H]
\centering
\includegraphics[width=\linewidth]{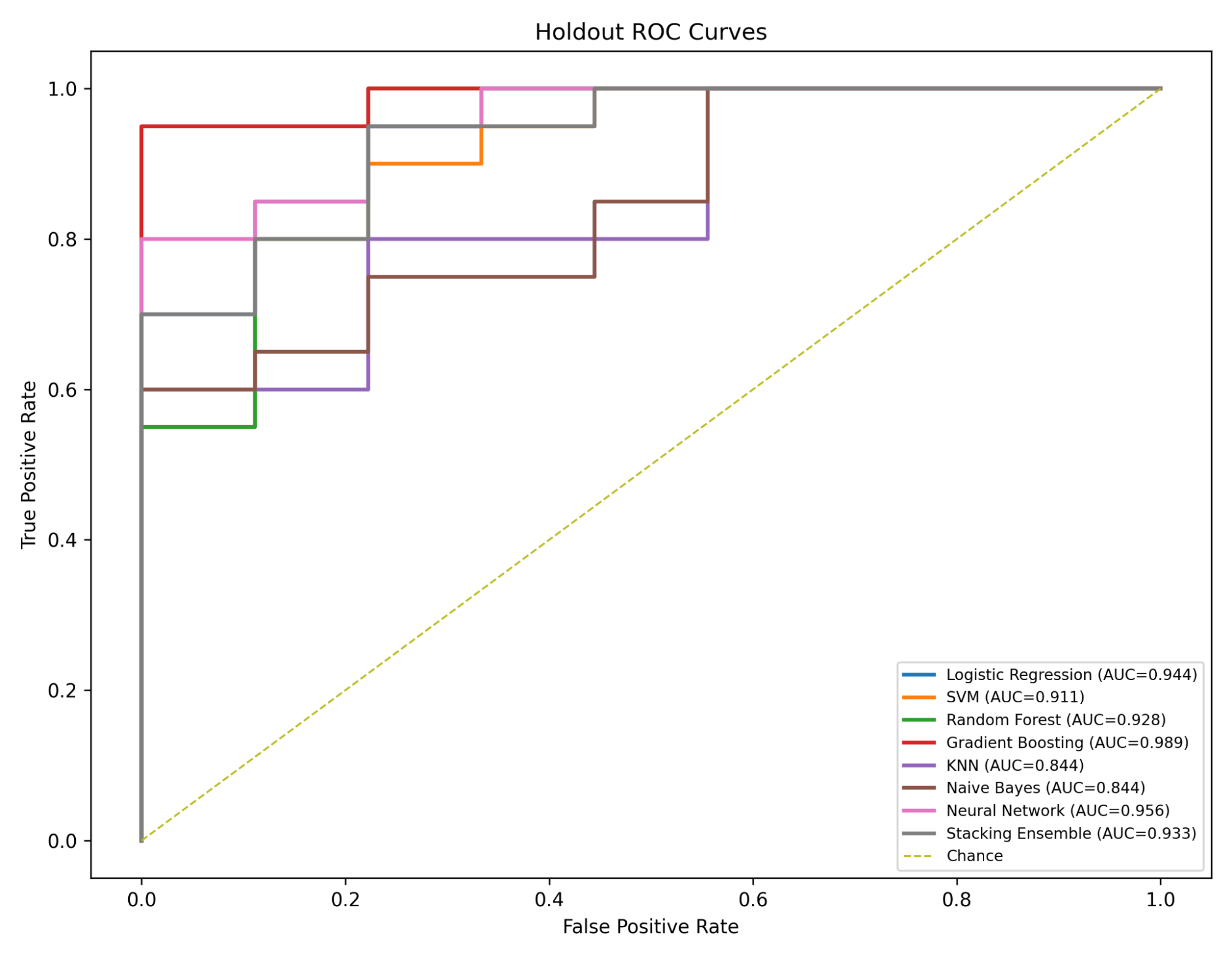}
\caption{Receiver operating characteristic (ROC) curves on the stratified holdout set. The plot displays the true positive rate against the false positive rate across various classification thresholds for the benchmark models, visualizing the high discrimination capacity (AUC-ROC) discussed in Section~III-C.}
\label{fig:roc}
\end{figure}

\subsection{Error Tradeoffs Across Models}
The holdout leaderboard shows several different error patterns. Random Forest detected all MS samples in the holdout set, but it mislabeled more healthy controls. KNN protected the healthy control class in this split, but it missed many MS samples. SVM reached high sensitivity with lower specificity. These differences would be hidden if the paper reported only accuracy or AUC. The full metric table is therefore part of the benchmark design.

\begin{figure}[H]
\centering
\includegraphics[width=\linewidth]{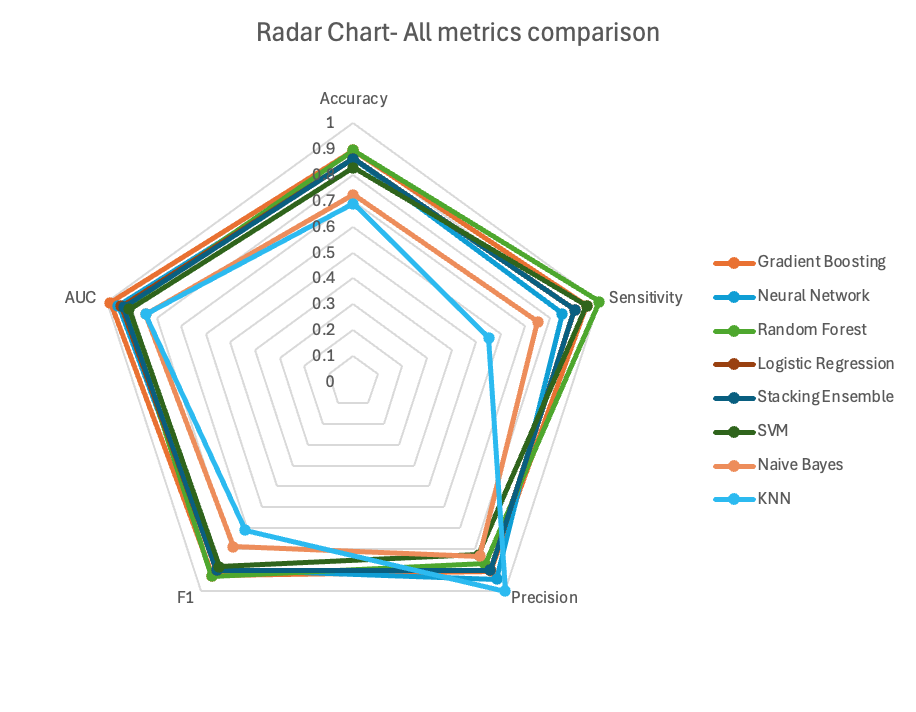}
\caption{Comparative radar plot of the machine learning models across AUC-ROC, recall, $F_1$-score, precision, and accuracy for whole-blood RNA-based multiple sclerosis detection. Gradient Boosting achieved the highest composite score, while the remaining leading models showed competitive performance with metric-specific tradeoffs. AUC-ROC values are scaled by 100 for visual consistency.}
\label{fig:radar}
\end{figure}

\subsection{Calibration and Probability Error}
Gradient Boosting also had the lowest Brier score in the stored holdout output. This contributed to its MS Research Score because calibration is part of the formula. Neural Network had the second highest score and a Brier score of 0.090. Random Forest had a higher $F_1$ score than Gradient Boosting, but its Brier score was 0.148 and its specificity was lower. These details explain why Gradient Boosting ranked first by the composite score.

\subsection{External Model Pathway}
The external model route remains part of MS-MLB, but the leaderboard reported in this paper focuses on the internal baseline models from the stored run. Future submissions can be evaluated against the same split, preprocessing pathway, metrics, and MS Research Score, so that outside estimators can be compared without rewriting the evaluation system.

\begin{table}[H]
\caption{Benchmark performance of machine learning models for whole-blood RNA-based multiple sclerosis detection. Models are ranked by the MS Research Score, with Gradient Boosting achieving the highest overall score. The table reports AUC, sensitivity, specificity, $F_1$-score, and Brier score to compare discrimination, classification balance, and calibration across models.}
\label{tab:leaderboard}
\centering
\small
\setlength{\tabcolsep}{3.5pt}
\begin{tabular}{lcccccc}
\toprule
\textbf{Model} & \textbf{Score} & \textbf{AUC} & \textbf{Sens.} & \textbf{Spec.} & \textbf{$F_1$} & \textbf{Brier} \\
\midrule
Gradient Boosting & 93.83 & 0.989 & 0.950 & 0.778 & 0.927 & 0.050 \\
Neural Network & 91.24 & 0.956 & 0.850 & 0.889 & 0.895 & 0.090 \\
Random Forest & 90.47 & 0.928 & 1.000 & 0.667 & 0.930 & 0.148 \\
Logistic Regression & 90.26 & 0.944 & 0.900 & 0.778 & 0.900 & 0.114 \\
Stacking Ensemble & 89.51 & 0.933 & 0.900 & 0.778 & 0.900 & 0.154 \\
SVM & 87.01 & 0.911 & 0.950 & 0.556 & 0.884 & 0.105 \\
Naive Bayes & 79.08 & 0.844 & 0.750 & 0.667 & 0.789 & 0.268 \\
KNN & 78.89 & 0.844 & 0.550 & 1.000 & 0.710 & 0.207 \\
\bottomrule
\end{tabular}
\end{table}

\begin{figure}[H]
\centering
\includegraphics[width=\linewidth]{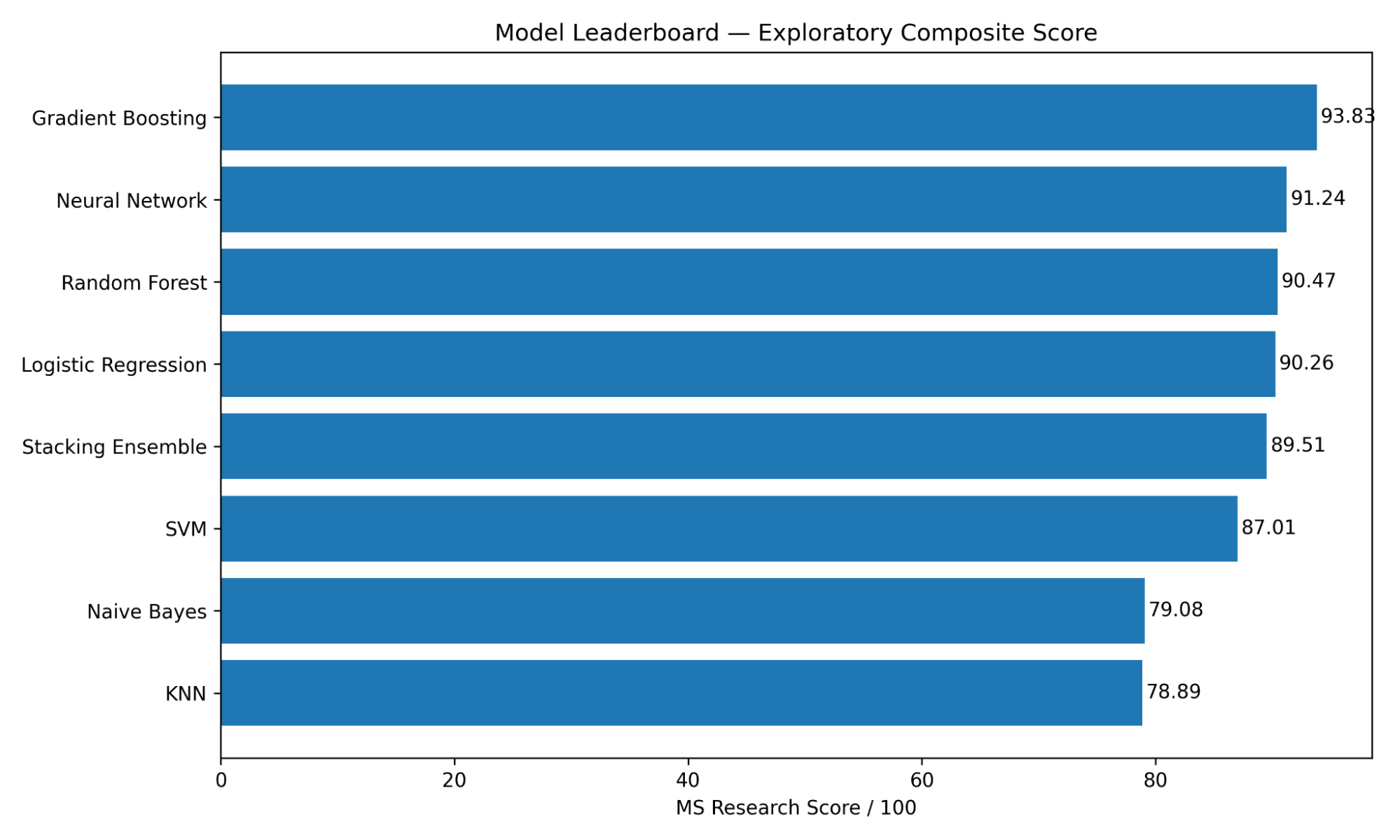}
\caption{Model leaderboard ranked by exploratory MS Research Score. The horizontal bar chart displays the benchmark algorithms ordered by their final composite MS Research Score evaluated on the stratified holdout set. Gradient Boosting achieves the highest baseline score at 93.83, followed by the Neural Network (91.24) and Random Forest (90.47). This visual ranking corresponds directly to the composite metrics outlined in the manuscript's primary results. Weighted: AUC-ROC (25\%) + PR-AUC (15\%) + sensitivity (25\%) + specificity (15\%) + $F_1$ (10\%) + calibration, measured as $1$ minus Brier score (10\%).}
\label{fig:leaderboard}
\end{figure}

\section{Analysis}

\subsection{Main Interpretation}
The results suggest that MS versus healthy control classification from GSE17048 whole blood RNA expression data is learnable under the benchmark protocol. Several models reached high AUC-ROC values on the holdout split. Gradient Boosting produced the strongest stored holdout result because it combined high discrimination, high sensitivity, acceptable specificity, strong $F_1$ performance, and low probability error.

The result is still bounded by the dataset. The holdout set contains 29 samples. A change of a few predictions would change sensitivity, specificity, $F_1$ score, and Brier score. The AUC confidence interval for Gradient Boosting was high, but it was estimated from a small test set. External validation is required before any biological or clinical claim can move beyond the benchmark.

\subsection{Why the Composite Score Helped}
The leaderboard shows why a single metric is insufficient. Random Forest reached a sensitivity of 1.000, but its specificity was 0.667. KNN reached a specificity of 1.000, but its sensitivity was 0.550. These models make different errors. The MS Research Score made those tradeoffs visible through one ranking while leaving the raw metrics available for inspection.

The score should be treated as a research summary. It is helpful for comparing model behavior inside this benchmark. It is not a diagnostic index. A future clinical setting may weight specificity, calibration, or threshold stability more heavily. Clinician input and external validation would be required before the weights could be used for decision support.

\subsection{Novelty of the Benchmark}
The paper makes a careful first-of-its-kind claim. The claim is limited to the open benchmark structure, since prior machine learning work on MS blood RNA already exists. Acquaviva et al. built PBMC transcriptomic classifiers for MS stages~\cite{acquaviva2020}. Omrani et al. evaluated whole blood transcriptomic signatures for MS and clinically isolated syndrome tasks~\cite{omrani2024}. Delkhah et al. integrated transcriptomic datasets and reported machine learning biomarker findings with limited generalization in external datasets~\cite{delkhah2025}.

The novelty here is the open benchmark structure. To our knowledge, MS-MLB is the first framework to cast MS biomarker ML detection from GSE17048 whole blood RNA as a reusable benchmark with fixed leakage controlled preprocessing, nested cross-validation, holdout evaluation, bootstrap intervals, calibration reporting, a defined MS Research Score, saved outputs, and a documented external model submission interface. This claim is narrower than saying the paper is the first MS biomarker ML paper. The narrower claim is more accurate and more useful.

\subsection{Model Behavior}
Gradient Boosting may have benefited from its ability to model nonlinear relationships after the feature selection step. That flexibility can help in gene expression data, where expression patterns may interact across immune pathways. It can also increase variance. The nested cross-validation scores should therefore stay beside the holdout scores in any presentation of the result.

Logistic Regression performed strongly despite its simpler structure. That result is scientifically useful. A linear model with stable selected genes may be easier to inspect than a higher scoring model whose decision path is harder to summarize. If future versions add feature stability analysis, Logistic Regression may serve as a useful comparator for biological interpretation.

KNN performed poorly on sensitivity, even though it achieved perfect specificity on the holdout split. Distance based methods often struggle in high dimensional settings because many features can add noise to sample distances. Naive Bayes also ranked low, which may reflect the mismatch between its feature independence assumption and correlated gene expression data.

\begin{figure}[!t]
\centering
\includegraphics[width=\linewidth]{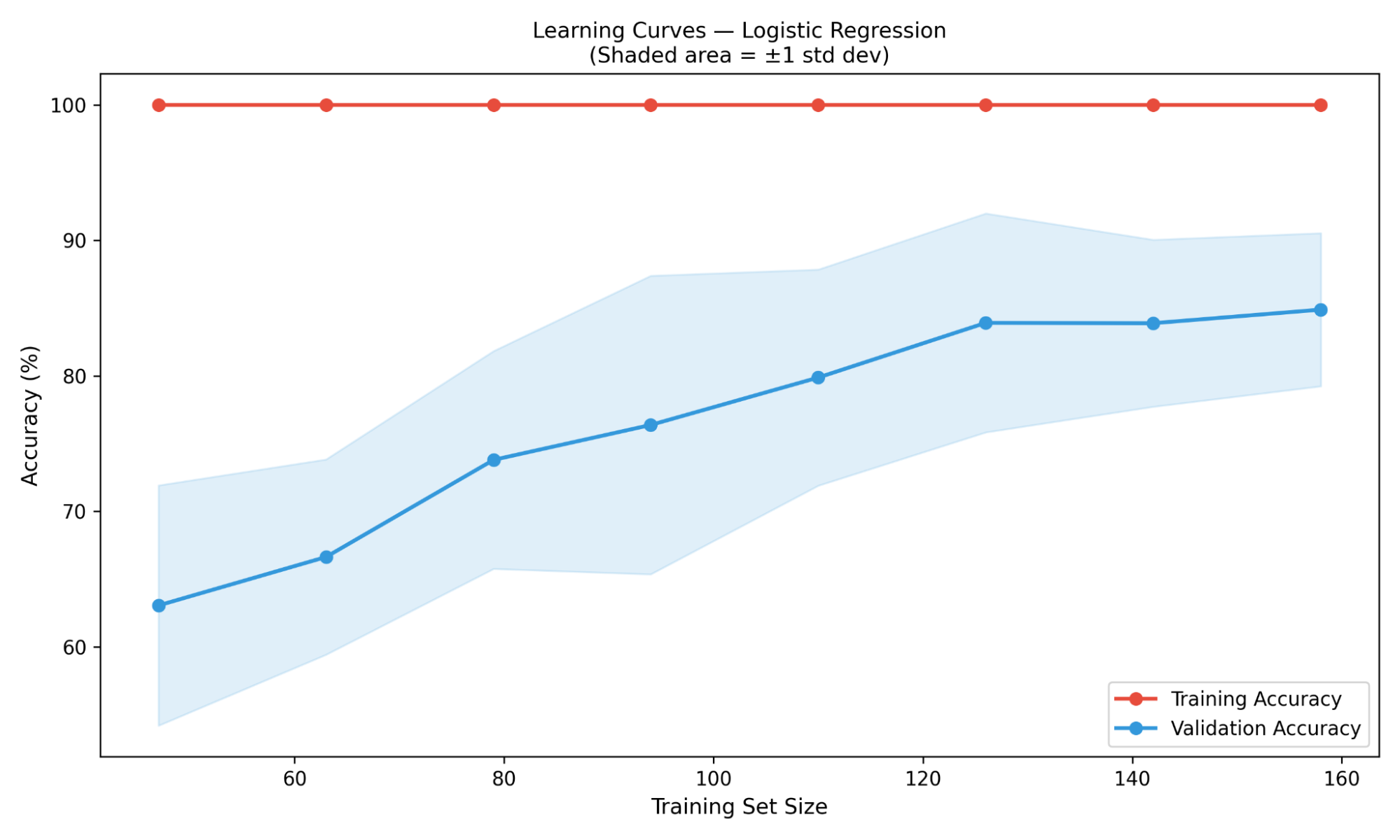}
\caption{Logistic regression learning curves across training set sizes. The line plot tracks model accuracy as a function of the training sample size, comparing performance on the training subset (red) against the validation subset (blue). The shaded region represents $\pm 1$ standard deviation around the cross-validation accuracy, illustrating how the model adapts to data volume and highlighting sample size constraints within the GSE17048 cohort.}
\label{fig:learning_curves}
\end{figure}

\subsection{Biological Interpretation}
The current benchmark is performance focused. It does not yet identify a validated gene signature. SelectKBest chooses features inside folds, but this paper does not report which genes were repeatedly selected. A future version should record selected features across folds and bootstrap samples, then test whether stable genes map to immune processes already reported in MS transcriptomic literature.

A predictive feature is not automatically a biomarker. A feature may represent disease biology, medication exposure, batch structure, subtype mixture, or another cohort feature. This is a major reason for keeping the wording conservative. The benchmark detects class signals in GSE17048. It does not prove that the selected RNA features are clinically usable biomarkers.

\begin{figure}[H]
\centering
\includegraphics[width=\linewidth]{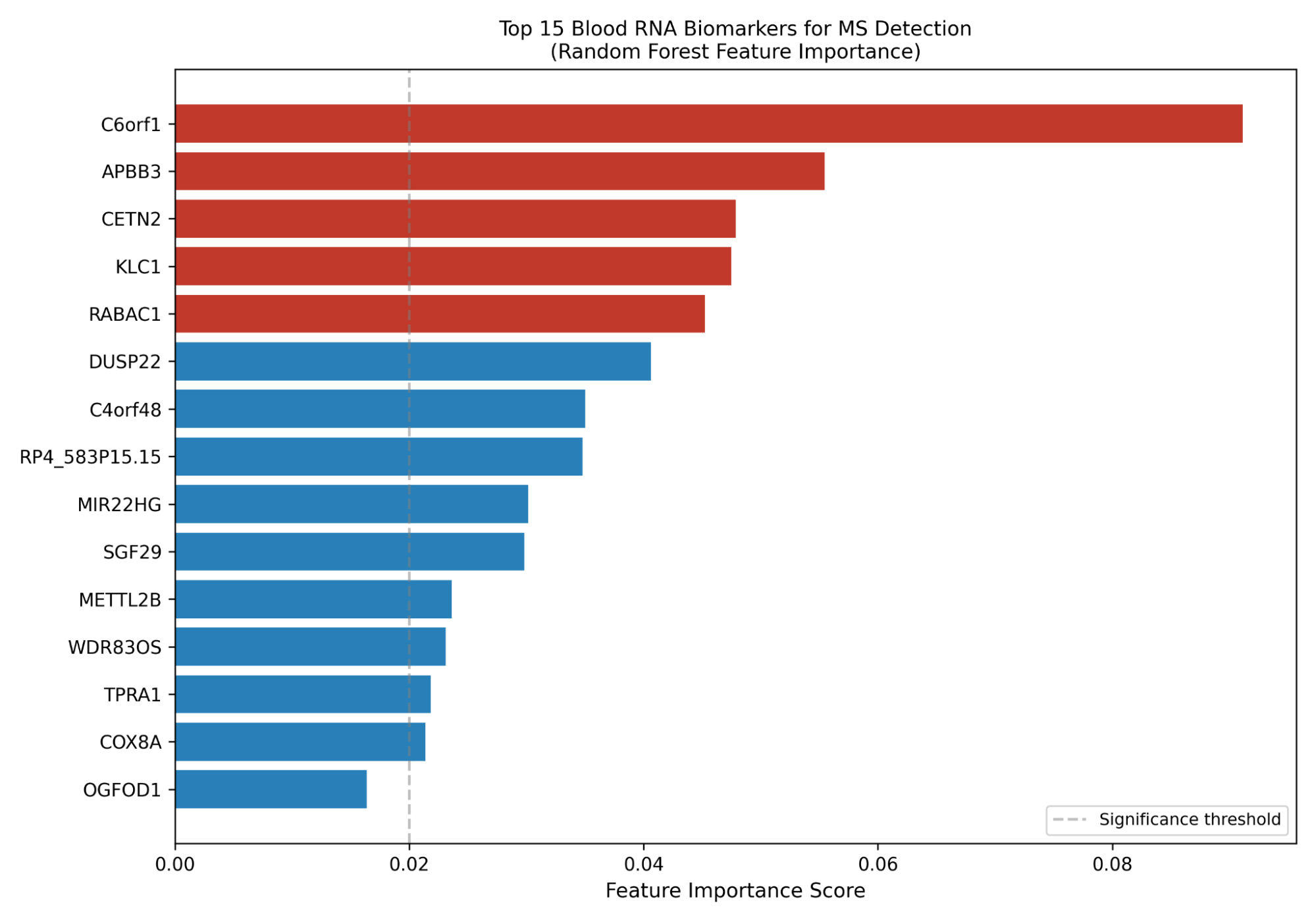}
\caption{Top 15 whole-blood RNA biomarkers ranked by Random Forest feature importance for multiple sclerosis detection. C6orf1 shows the highest importance score, followed by APBB3, CETN2, KLC1, and RABAC1. Red bars indicate biomarkers exceeding the predefined significance threshold, while blue bars represent additional ranked features contributing to model classification performance.}
\label{fig:biomarkers}
\end{figure}

\subsection{Generalization and Confounding}
Single cohort modeling is vulnerable to confounding. Age, sex, treatment status, blood draw conditions, RNA processing, array batch, and site effects can influence gene expression. If any of these variables correlate with MS status, a classifier may use them indirectly. The current public dataset does not allow full correction for every possible confounder. External validation across independent cohorts is the best next test.

That limitation explains why the benchmark is needed. A reusable framework can expose whether a model still performs when the same evaluation rules are applied to a new dataset. If future external validation shows weaker performance, the benchmark will still have served its purpose by identifying the gap between internal performance and generalization.

\subsection{External Model Value}
The external model pathway is the main reason MS-MLB should be treated as a benchmark paper rather than a single model report. A reader can add an estimator and rerun the comparison against the same split, pipeline, metrics, and score. This turns the benchmark into a shared testing system rather than a fixed leaderboard, and it means the effort a new user spends is on their model, not on rebuilding the evaluation around it.

Pretrained models need stricter handling. A pretrained model can only be compared fairly if the training data, feature order, preprocessing, and prior exposure to the benchmark dataset are disclosed. The framework should eventually separate tunable external models from pretrained external models in the result table. That separation would make the leaderboard easier for a reader to interpret at a glance.

\subsection{Comparison with Prior MS Transcriptomic ML Work}
The prior literature makes the novelty claim more precise. Acquaviva et al. used PBMC transcriptomic profiles to infer MS stages~\cite{acquaviva2020}. Omrani et al. moved to whole blood transcriptomics and reported machine learning based MS classification and CIS related analyses~\cite{omrani2024}. Delkhah et al. integrated PBMC expression datasets and reported biomarker candidates with machine learning, while also noting generalization limits in external datasets~\cite{delkhah2025}. These papers support the biological and methodological background for this research.

The present paper occupies a different position. Its main result is a benchmark that can be rerun, inspected, and extended. The difference matters because reproducibility is a known challenge in small high dimensional biomedical ML. A benchmark can be useful even if later models replace the current top performer. The protocol, saved outputs, and external submission route remain usable.

\subsection{Why the First-of-Its-Kind Claim Is Narrow}
The phrase first-of-its-kind can be risky in biomedical writing. Here it is used in a narrow way. The paper limits the statement to the first open reusable benchmark paper for GSE17048 whole blood RNA MS versus healthy control classification with external model testing as part of the framework, to our knowledge. That wording matches the evidence more closely.

\subsection{Clinical Boundary}
The clinical boundary should remain visible throughout the paper. MS diagnosis requires clinical context, imaging evidence, and exclusion of better explanations~\cite{thompson2018}. A benchmark trained on public blood expression data cannot evaluate mimic disorders, real clinic referral populations, treatment decisions, or safety consequences. The score is named MS Research Score for that reason. A clinical diagnostic score would need a different development process and independent validation.

\subsection{Data and Platform Limitations}
GSE17048 is based on microarray expression data. Modern RNA sequencing datasets may have different feature distributions, measurement noise, and preprocessing requirements. A future version that combines microarray and RNA sequencing sources will need a careful gene mapping and normalization plan. Directly applying a model trained on one platform to another platform can fail if feature definitions or distributions do not match.

\subsection{Reporting Quality}
The paper also benefits from reporting negative space clearly. It states what the benchmark does not test: subtype classification, CIS conversion, treatment response, prognosis, and clinical diagnosis. This makes the claim stronger because the task is defined precisely. A narrow task with reproducible code is easier to evaluate than a broad claim with unclear boundaries.

\section{Conclusion}
This paper presents MS-MLB, a reproducible open benchmark for machine learning based MS classification from whole blood RNA expression data. The framework uses the public GSE17048 cohort, compares multiple algorithms under one evaluation pathway, controls preprocessing leakage through pipeline placement, reports nested cross-validation and holdout testing, calculates bootstrap confidence intervals, includes calibration measurement, and ranks models with an exploratory MS Research Score.

In the stored benchmark run, Gradient Boosting produced the strongest holdout result, followed by Neural Network, Random Forest, Logistic Regression, and Stacking Ensemble. The results support the presence of learnable MS associated signal in the dataset. They should be read as benchmark evidence within one public cohort.

The strongest contribution is the first open benchmark framing for this specific MS whole blood RNA classification task, to our knowledge. Earlier papers have used machine learning on MS blood transcriptomics. This paper defines a reusable model comparison system with external model testing. That distinction keeps the novelty claim accurate.

Clinical diagnostic use is outside the evidence presented here. Larger external cohorts, feature stability analysis, pathway mapping, calibration assessment, and score sensitivity testing are needed before stronger claims can be made.

MS-MLB is therefore best presented as a benchmark infrastructure paper with a strong stored baseline. The first result is the trained model comparison. The second result is the reusable evaluation route that other researchers can operate with minimal setup. The second result is the reason the work can be extended by other researchers rather than ending with the current leaderboard.

\begin{figure}[H]
\centering
\includegraphics[width=\linewidth]{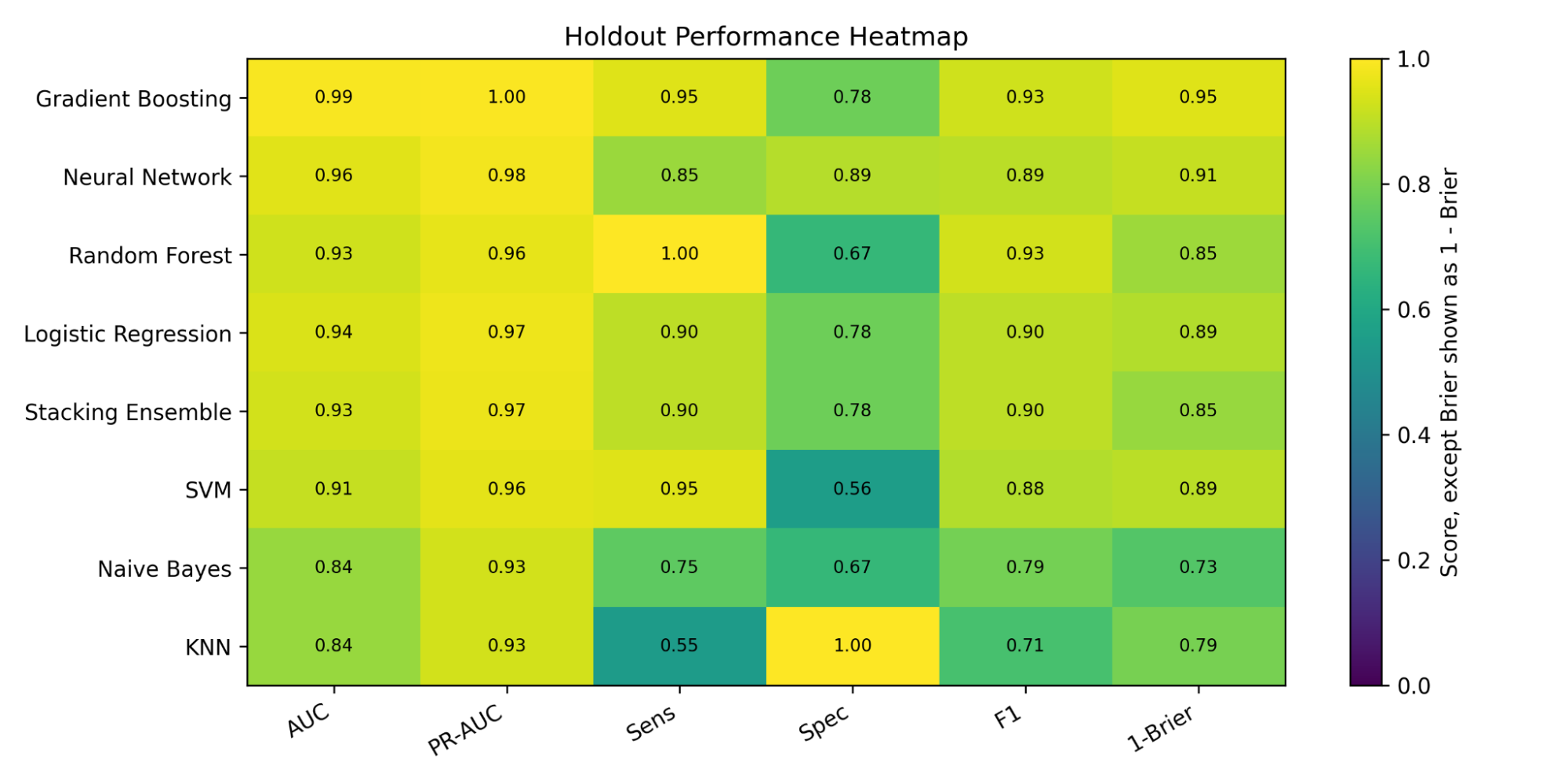}
\caption{Methodological pipeline for next-generation benchmark scaling. This structural flowchart maps the ten sequential technical enhancements outlined in Sections~VI-A through VI-J. The framework partitions future development into empirical refinement validation phases (left) and repository reproducibility governance standards (right), serving as an engineering blueprint to transition the repository from a single-cohort evaluation system to an audited, multi-platform machine learning registry.}
\label{fig:pipeline}
\end{figure}

\section{Improvements}

\subsection{Independent Validation}
The first improvement is external validation. Future work should test the benchmark on additional whole blood RNA datasets collected by different groups and platforms. This would show whether the models are learning MS associated expression patterns or cohort specific structure. If feature overlap differs across platforms, a gene mapping rule should be defined before training.

\subsection{Feature Stability}
The second improvement is feature stability analysis. The benchmark should save the genes selected in every fold and every bootstrap repetition. Genes selected repeatedly can then be compared with known immune biology and prior MS transcriptomic findings. This would make the benchmark more useful for biomarker discovery.

\subsection{Biological Pathway Analysis}
The third improvement is pathway analysis. Stable feature sets can be mapped to immune pathways, interferon response, T cell regulation, oxidative stress, or other pathways reported in MS blood expression work. This should be treated as hypothesis generation. The paper should avoid presenting selected genes as validated biomarkers unless they replicate across independent cohorts.

\subsection{Calibration and Thresholds}
The fourth improvement is a fuller calibration module. The current benchmark reports Brier score. Later versions should include calibration curves, calibration slope, calibration intercept, and threshold dependent performance. The default threshold of 0.5 is simple, but it may not match a research use case that prioritizes sensitivity or specificity.

\subsection{Score Sensitivity}
The fifth improvement is score sensitivity analysis. The MS Research Score should be recalculated with alternative weights. If Gradient Boosting remains near the top across plausible scoring schemes, the ranking becomes more stable. If the ranking changes sharply, the paper should state that the leaderboard depends on the chosen weights.

\subsection{External Model Governance}
The sixth improvement is a stronger submission standard. External models should include a short model card, feature order requirements, package versions, training data disclosure, preprocessing details, and a statement about prior exposure to GSE17048. The benchmark should automatically warn users when a pretrained model cannot be compared fairly.

\subsection{Repository Cleanup}
The repository should keep terminology consistent. The paper uses MS Research Score because the score is exploratory. The README should avoid language that suggests clinical validation. The score weights in the README should also match the metadata and code. This will make the repository and paper read as one aligned research.

\subsection{Dataset Expansion Plan}
A practical next step is to add a dataset registry. Each dataset should include source, platform, sample count, class labels, available covariates, preprocessing status, and whether it is used for training, validation, or external testing. This would prevent accidental reuse of the same cohort across development and validation. It would also make the benchmark easier to audit.

\subsection{Dependency Locking}
The repository should add a locked requirements file created from the exact environment used for the stored run. The current metadata records the Python version and states that exact package versions can be saved with pip freeze~\cite{simson2026}. A lock file would make the result easier to reproduce on another machine. It would also help detect changes caused by package updates.

\subsection{Model Cards and Submission Checks}
External submissions should use a short model card. The card should state the model type, training data, preprocessing, feature order, package versions, and any exposure to the benchmark dataset. The benchmark can then label each external result as fold-trained, holdout-only, or pretrained. This separation would protect the leaderboard from unfair comparisons.

\balance

\end{document}